# Graphic Symbol Recognition using Graph Based Signature and Bayesian Network Classifier

Muhammad Muzzamil Luqman, Thierry Brouard and Jean-Yves Ramel
*Université François Rabelais de Tours, Laboratoire d'Informatique (EA 2101)*
*64, Avenue Jean Portalis, 37200 Tours – France*
muhammadmuzzamil.luqman@etu.univ-tours.fr, {brouard, ramel}@univ-tours.fr

**Abstract**

*We present a new approach for recognition of complex graphic symbols in technical documents. Graphic symbol recognition is a well known challenge in the field of document image analysis and is at heart of most graphic recognition systems. Our method uses structural approach for symbol representation and statistical classifier for symbol recognition. In our system we represent symbols by their graph based signatures: a graphic symbol is vectorized and is converted to an attributed relational graph, which is used for computing a feature vector for the symbol. This signature corresponds to geometry and topology of the symbol. We learn a Bayesian network to encode joint probability distribution of symbol signatures and use it in a supervised learning scenario for graphic symbol recognition. We have evaluated our method on synthetically deformed and degraded images of pre-segmented 2D architectural and electronic symbols from GREC databases and have obtained encouraging recognition rates.*

## 1. Introduction and related works

Graphics recognition is a subfield of document image analysis and it deals with graphic entities that appear in document images. As pointed out by Lladós and Sánchez in [*1*]: documents from electronics, engineering, music, architecture and various other fields use domain-dependent graphic notations which are based on particular alphabets of symbols. These industries have a rich heritage of hand-drawn documents and because of high demands of application domains, overtime symbol recognition is becoming core goal of automatic image analysis systems. Some typical applications of symbol recognition include hand-drawn based user interfaces, backward conversion from raster images to CAD, content based retrieval from graphic document databases and browsing of graphic documents. A detailed discussion on application domains is in [*2*, *3*] and a quick historical overview of the work on graphic symbol recognition is given by Tombre et al. [*4*].

Graphic symbol recognition is generally approached by structural methods of pattern recognition which normally use graph based representations and thus inherit the various advantages associated with these representations. These methods, for example [*5*, *6*] and the methods mentioned in [*1*], then employ graph matching or graph comparison techniques for symbol recognition. Graph matching and graph comparison are time consuming tasks and they limit the ability of these systems to scale to large number of symbol models. Moreover, structural methods generally require in-depth domain knowledge and this hinders the possibility of having a generalized system of symbol recognition. Another approach for graphic symbol recognition is use of statistical methods of pattern recognition. These methods represent graphic symbol by feature vector or signature (we use these terms interchangeably) and use a statistical classifier for symbol recognition. The use of signatures and statistical classifiers allows designing of fast and efficient systems which are sufficiently scalable and domain independent. A state of the art for various methods that employ different structural or statistical approaches for graphic symbol recognition is in [*7*].

The rest of paper is organized as follows: section 2 is devoted to general description of our method and section 3 provides detailed description of each part of system. Experimental results are presented in section 4 and we present some concluding remarks and future directions of work in section 5.

## 2. Proposed method

### 2.1 A combination of structural and statistical approaches

We have approached the problem of graphic symbol recognition by employing a structural method for symbol representation and a statistical classifier for recognition. In this paper we take forward the work of Qureshi et al. [8]. They vectorize a graphic symbol, construct its attributed relational graph and compute a structural signature (G-signature as they call it). For classification of query symbol they use nearest neighbor rule with Euclidian distance as measure of dissimilarity. The structural signature is discriminant in case of hand-drawn or vectorial deformations and has been shown invariant of rotation and scaling. We argue that the computation of Euclidian distance in a brute force manner (between query symbol and each prototype in training set) limits this system to scale to large number of symbol models or to be used by real time applications. The system is based on vectorization and faces a high degree of uncertainty as the level of noise and deformation increase. In our system we use structural signature with a statistical classifier. We have selected Bayesian networks for dealing with uncertainty in symbol signatures. We deal only with linear graphic symbols in this work i.e. symbols that consist of only straight lines and arcs. This gives us a chance to optimize the structural signature for these types of symbols. The signature is given in Figure 3 and it is discussed in section 3.2.

## 2.2 Bayesian networks

Bayesian networks are probabilistic graphical models and are represented by their structure and parameters. Structure is given by a directed acyclic graph and it encodes the dependency relationships between domain variables whereas parameters of the network are conditional probability distributions which are associated with its nodes. A Bayesian network like other probabilistic graphical models encodes joint probability distribution of a set of random variables and could be used to answer all possible inference queries on these variables. A humble introduction to Bayesian networks is in [9] and [10].

Bayesian networks have already been applied successfully to a large number of problems in machine learning and pattern recognition and are well known for their power and potential of making valid predictions under uncertain situations. But in our knowledge there are only a few methods which use Bayesian networks for graphic symbol recognition. Recently Barrat et al. [11] have used the naïve bayes classifier in a 'pure' statistical manner for graphic symbol recognition. Their system use three shape descriptors (Generic Fourier Descriptor, Zernike descriptor and R-Signature 1D) and applies dimensionality reduction for extracting the most relevant and discriminating features to formulate a feature vector. This reduces the length of their feature vector and eventually the number of variables (nodes) in network. The naïve bayes classifier is a powerful Bayesian classifier but it assumes a strong independence relationship among attributes given class variable. We believe that the power of Bayesian networks is not fully explored; as instead of using pre-defined dependency relationships we can obtain a better Bayesian network classifier if we find dependencies between all variable pairs from underlying data.

## 2.3 Originality of our approach

Our method is an original adaptation of Bayesian network learning for the problem of graphic symbol recognition. We use a structural signature for symbol representation. The signature is computed from the attributed relational graph of graphic symbol and is composed of geometric and topologic characteristics of the structure of symbol. We use a Bayesian network for symbol recognition. This network is learned from underlying training data by using the quite recently proposed genetic algorithms for Bayesian network learning by Delaplace et al. [12]. A query symbol is classified by using Bayesian probabilistic inference (on encoded joint probability distribution).We have selected the features in signature very carefully to best suit them to linear graphic symbols and to restrict their number to minimum; as Bayesian network algorithms are known to perform better for a smaller number of nodes. The use of structural signature makes our system independent of application domains and it could be used for all types of 2D linear graphic symbols. Also, relatively basic computations are involved for recognizing a query symbol which enables our system to respond in real time and it could be used for instance as a pre-processing step of a traditional symbol recognition method or for indexation and browsing of graphic documents.

## 3. Detailed description

Cordella and Vento [7] have remarked that a graphics recognition system can be looked upon as working in three phases: representation phase, description phase and classification phase. In this section we describe our system in light of these phases.

### 3.1 Representation phase

As we have stated earlier; our method of graphic symbol recognition is continuation of work already done and the representation phase of our method is exactly same as that of [8]. Figure 1 outlines different steps that are involved in representation phase. A graphic symbol is vectorized and is represented by a set of primitives (quadrilaterals & vectors). Thin parts of shape are represented by quadrilaterals and filled regions by vectors. Our system deals only with linear graphic symbols and hence all our symbols are composed of only thin regions which are represented by quadrilaterals. The vectorization is followed by construction of an attributed relational graph whose nodes are graphic primitives (quadrilaterals) and arcs show relationships of connectivity between them.

### 3.2 Description phase

We use the attributed relational graph produced in representation phase for computing structural signature for symbol. The signature exploits the structural information and encodes structural details of a symbol; in order to differentiate it from pixel based or statistical signatures we call it a structural signature. To best suit the feature vector to the type of symbols that we deal with, we propose a different set of features than [8]. Our signature does not contain features concerned with primitive of type 'vector'. We normalize the relative angle between 0° and 90° and use different set of length and angle intervals for computing range features (Figure 2). The list of 21 features in our signature of graphic symbol is given in Figure 3. Quantitative features in signature encode details of the size of symbol and density of connections at its primitives. Symbolic features encode the details about shape of symbol and help to discriminate between symbols of similar size (number of primitives) but different shape (arrangement of primitives). And the range features exploit the attributes of the primitives and serve as complementary criteria for discriminating between different symbol classes.

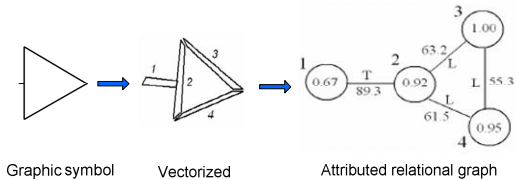

**Figure 1.** Representation phase.

| Relative length | Relative angle |
|---|---|
| Length1 : 0.00 – 0.19 | Angle1 :  0° – 29° |
| Length2 : 0.20 – 0.39 | Angle2 : 30° – 59° |
| Length3 : 0.40 – 0.59 | Angle3 : 60° – 90° |
| Length4 : 0.60 – 0.79 | |
| Length5 : 0.80 – 1.00 | |

**Figure 2.** Relative length and angle intervals.

| | |
|---|---|
| Quantitative Features | f1  : Number of nodes |
| | f2  : Number of Arcs |
| | f3  : Number of nodes connected to 1 node |
| | f4  : Number of nodes connected to 2 nodes |
| | f5  : Number of nodes connected to 3 nodes |
| | f6  : Number of nodes connected to 4 nodes |
| | f7  : Number of nodes connected to 5 nodes |
| | f8  : Number of nodes connected to 6(+) nodes |
| Symbolic Features | f9  : Number of arcs with label 'L' |
| | f10 : Number of arcs with label 'P' |
| | f11 : Number of arcs with label 'T' |
| | f12 : Number of arcs with label 'X' |
| | f13 : Number of arcs with label 'S' |
| Range Features | f14 : Number of nodes in interval 'Length1' |
| | f15 : Number of nodes in interval 'Length2' |
| | f16 : Number of nodes in interval 'Length3' |
| | f17 : Number of nodes in interval 'Length4' |
| | f18 : Number of nodes in interval 'Length5' |
| | f19 : Number of arcs in interval 'Angle1' |
| | f20 : Number of arcs in interval 'Angle2' |
| | f21 : Number of arcs in interval 'Angle3' |

**Figure 3.** List of features in signature.

### 3.3 Learning and classification phase

In our system graphic symbols are represented by their signatures. We discretize our learning and test datasets because the Bayesian network algorithms, which we have used, require discrete data. We learn a Bayesian network from the discretized learning data and use it in a supervised learning context for assigning labels to signatures of unknown query symbols (from discretized test data).

**3.3.1 Discretization.** We achieve discretization or quantification of datasets by a histogram based technique which is available in Bayesian Network Structure Learning Package of François and Leray [13]. This technique is based on use of Akaike Information Criterion (AIC). It starts with an initial m-bin histogram of data and finds optimal number of bins for underlying data. Two adjacent bins are merged using an AIC-based cost function as criterion; until the difference between AIC-before-merge and AIC-after-merge becomes negative. Each row of Figure 4 corresponds to a feature vector (21 feature variables plus class variable) and each column can be looked upon as probability distribution of a variable. We discretize each variable separately and independently of other variables. The class labels are chosen intelligently in order to avoid the need of any discretization for them.

```
8,11,0,5,1,1,1,0,6,0,3,0,2,1,0,3,2,2,2,1,8,002
8,11,0,5,1,1,1,0,6,0,3,0,2,1,0,3,2,2,2,1,8,002
4,5,0,2,2,0,0,0,2,0,3,0,0,0,0,0,0,4,0,1,4,016
6,9,0,2,2,2,0,0,7,0,0,0,2,0,0,4,0,2,2,3,4,016
```

**Figure 4.** A snapshot of learning data containing signatures with class labels.

**3.3.2 Learning step.** We learn the Bayesian network in two stages; namely structure learning stage and parameter learning stage. Goal of structure learning stage is to find the best network structure from underlying data which contains all possible dependency relationships between all variable pairs. This is achieved by genetic algorithms of Delaplace et al. [12]. Figure 5 shows one of the learned structures from our experiments (each node corresponds to a feature variable). The parameters of network are conditional probability distributions which are associated with its nodes and they specify conditional probability of the node given probabilities of its parents. The network parameters are obtained by maximum likelihood estimation (MLE); which is a robust parameter estimation technique and assigns the most likely parameter values to best describe a given distribution of data. We make use of Dirichlet priors with MLE to avoid null probabilities. The learned Bayesian network encodes joint probability distribution of symbol signatures in learning dataset.

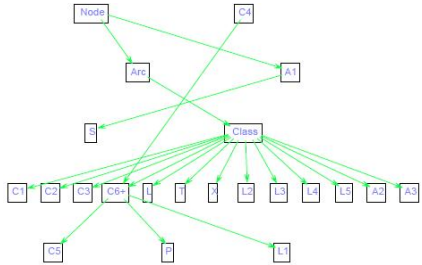

**Figure 5.** A Bayesian network structure after learning step; each node corresponds to a feature variable.

**3.3.3 Classification step.** Bayesian probabilistic inference on encoded joint probability distribution is deployed for assigning a class label to signature of query symbol. The idea behind this is to compute the list of probabilities with which each class label can be assigned to a query symbol. Probabilistic inference is achieved by using junction tree inference engine which is the most popular exact inference engine for Bayesian network probabilistic inference and is available in [13]. The inference engine propagates the evidence (signature of query symbol) in network and computes posterior probability for each class label. This in fact refers to Bayes rule which is given below by Eq. (1).

$$P(c_i | e) = \frac{P(e, c_i)}{P(e)} = \frac{P(e | c_i) \times P(c_i)}{P(e)} \quad (1)$$

Where,

$$e = f1, f2, ..., f21$$

$$P(e) = P(e, c_i) = \sum_{i=1}^{k} P(e | c_i) \times P(c_i)$$

Eq. (1) states that posterior probability or probability of class *'ci'* given an evidence *'e'* is computed from likelihood (probability of evidence given class *'ci'*), prior probability of class *'ci'* and marginal likelihood (prior probability of evidence). After computing posterior probabilities for all class labels, we assign query signature to class which maximizes posterior probability i.e. which has highest posterior probability.

## 4. Experimental results

We have experimented with synthetically generated 2D symbols of models collected from databases of GREC symbol recognition contest [14]. There are a total of 150 models in GREC databases. In order to get a true picture of the performance of the proposed method on this database, we have randomly selected subsets with 100, 75, 50 and 20 different classes and generated our learning and test sets for each of these subsets. For each class the perfect symbol along with its 36 rotated and 12 scaled examples was used for learning; as the features have already been shown invariant to scaling and rotation [8] and because of the fact that generally Bayesian network learning algorithms perform better on datasets that contain quite a good number of examples. The system has been tested for its scalability on clean symbols (rotated/scaled), various levels of vectorial deformations (Figure 7) and binary degradations (Figure 6) of GREC symbol recognition context. Each test dataset contained 10 query symbols per class.

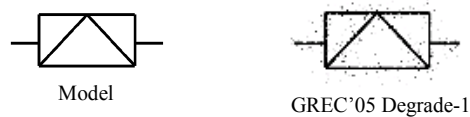

**Figure 6.** Model of degradation used to simulate photocopying / printing / scanning; applied using ImageMagick[1] and QGar package[2].

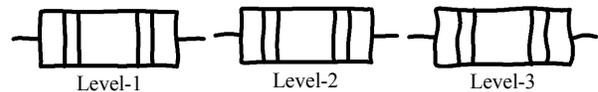

**Figure 7.** Models of deformation used for simulating hand-drawn symbols; applied using project Epéires[3].

---

[1] http://www.imagemagick.org/
[2] http://www.qgar.org/
[3] http://www.epeires.loria.fr/

**Table 1.** Symbol recognition experimental results.

| Number of classes | | 20 | 50 | 75 | 100 |
|---|---|---|---|---|---|
| **Clean symbols** | | 100% | 100% | 100% | 100% |
| Hand-drawn deformation | **Level-1** | 99% | 96% | 93% | 92% |
| | **Level-2** | 98% | 94% | 92% | 91% |
| | **Level-3** | 91% | 77% | 71% | 69% |
| **Binary degrade** | | 98% | 95% | 93% | 92% |

Table 1 summarizes the experimental results. A 100% recognition rate for clean symbols illustrates the invariance of our method to rotation and scaling. The recognition rates decrease with level of deformation and drop drastically for high binary degradations because of irregularities produced in symbol signature; which is a direct outcome of the noise sensitivity of vectorization step. We have not used any sophisticated de-noising or pretreatment and our method derives its ability to resist against noise, directly from underlying vectorization technique. Figure 8 gives a comparison of recognition rates. The system proposed in [8] presents recognition rates only for 20 models.

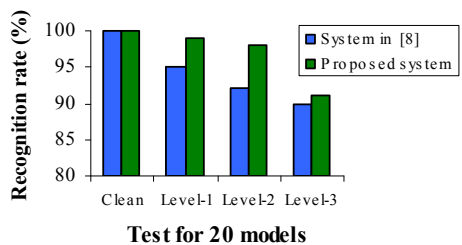

**Figure 8.** Comparison of recognition rates.

## 5. Conclusion

We have presented an original adaptation of Bayesian network learning for the problem of graphic symbol recognition. Our signature exploits the structural details of symbols. We represent symbols by signatures and encode their joint probability distribution by a Bayesian network. We then use Bayesian probabilistic inference on this network to classify query symbols. Experimental results of our method shows an improvement in recognition rates obtained by system in [8] and shows the scalability of the proposed system. Our system does not use any sophisticated de-noising/pretreatment and it drives its power to resist against deformations and degradations, directly from representation phase. The features in signature are affected by the small quadrilaterals that are produced during vectorization (in case of noisy symbols) and produces irregularities in signature. The use of Bayesian networks and Bayesian probabilistic inference gives our system a certain level of resistance against these irregularities. Our initial experiments have produced encouraging results and we have found that the system is scalable to sufficiently large number of models (classes) but with moderate levels of deformation and degradation. We believe that the recognition rates will be improved for real learning sets which include deformed and degraded examples as well. The system is extensible to new models and it has the ability to work for 2D linear graphic symbols from any domain. Offline learning and use of lightweight signature makes our system suitable for applications which involve indexation, retrieval and browsing of graphic documents. In future we plan to take this work forward to increase robustness of our signature against noise and deformations by introducing fuzzy intervals for computing quantitative features and range features.